\documentclass{article}
\usepackage{ijcai18}

\usepackage{times}
\usepackage{xcolor}
\usepackage{soul}
\usepackage[utf8]{inputenc}
\usepackage[small]{caption}

\usepackage{amssymb}
\usepackage{amsmath}
\usepackage{relsize}
\usepackage{algorithm,algorithmic}
\usepackage{graphicx}
\usepackage{url}
\usepackage{bbm}

\title{Accelerating Stochastic Gradient Descent Using Antithetic Sampling}

\author{
Jingchang Liu,
Linli Xu,
\\
University of Science and Technology of China \\%
xdjcl@mail.ustc.edu.cn,
linlixu@ustc.edu.cn
}

\begin{document}

\maketitle

\begin{abstract}
(Mini-batch) Stochastic Gradient Descent is a popular optimization method which has been applied to many machine learning applications. But a rather high variance introduced by the stochastic gradient in each step may slow down the convergence. In this paper, we propose the antithetic sampling strategy to reduce the variance by taking advantage of the internal structure in dataset. Under this new strategy, stochastic gradients in a mini-batch are no longer independent but negatively correlated as much as possible, while the mini-batch stochastic gradient is still an unbiased estimator of full gradient. For the binary classification problems, we just need to calculate the antithetic samples in advance, and reuse the result in each iteration, which is practical. Experiments are provided to confirm the effectiveness of the proposed method.
\end{abstract}

\section{Introduction}
In this big data era, many machine learning applications involve a large number of samples. To solve these problems effectively, Stochastic Gradient Descent~(SGD) is a desired method and  has been extensively studied in the machine learning community~\cite{Bottou2016,allen2017natasha,duchi2010composite,johnson2013accelerating,zhao2015stochastic}. In each iteration, a typical SGD will draw one sample from training data to calculate the stochastic gradient. The stochastic gradient is an unbiased estimator of full gradient. A nature modification is to employ the mini-batch training, which draws several samples to calculate the mini-batch stochastic gradient. And the mini-batch stochastic gradient is also unbiased. In the traditional mini-batch SGD setting, each sample is drew independently to compose the mini-batch, which will lead to a rather high variance. The variance may weaken the performance of SGD. In this paper, we introduce the antithetic sampling strategy to SGD to reduce the variance, while keeping the unbiased property of stochastic gradient.

In order to induce a smaller variance, the requirement of independence between every two samples inside a mini-batch is removed in our method. Exactly, if the two stochastic gradients derived from two samples are negatively correlated, then they would generate a mini-batch stochastic gradient with smaller variance. In other words, a smaller variance of this mini-batch stochastic gradient is foreseeable when the inner product of these two stochastic gradients being smaller in the expectation. We regard such two samples are antithetic to each other. The mutually antithetic samples compose the mini-batch in our method. Unlike the stratified sampling in SGD which needs a relatively big mini-batch~\cite{zhao2014accelerating}, the size of mini-batch required in the antithetic sampling is quiet flexible: it just need to be bigger than $1$.


To get the antithetic samples, one need to calculate the inner products between every two stochastic gradients in each step, which is inefficient. A more practical approach is to relax the problem: we turn to reduce the bound of the inner product to expect a smaller inner product. For the binary classification problem, the bound can be easily got, and it is independent of optimization variables. As a result, we just need to do calculations beforehand to get the antithetic table which records the information of antithetic samples, and reuse this table in each iteration. Since some multi-class classification problems can be split into several binary classification problems to solve~\cite{aly2005survey}, our method can be expanded to these multi-class classification problems. Besides, our method is complementary to other variance reduction methods~\cite{johnson2013accelerating,zhao2014accelerating,zhao2015stochastic}. In fact, different variance reduction methods can be combined to use.

The rest of this paper is organized as follows. Section 2 reviews the related work. In section 3, we will analyse the impacts of variance to convergence performance, and introduce the antithetic sampling strategy to SGD. In section 4, we relax the problem of calculating antithetic samples, and propose practical methods for the common binary classification problems. The empirical evaluations are presented in section 5. Section 6 concludes the paper.

\section{Related Work}

Stochastic Gradient Descent~(SGD) was studied long ago~\cite{robbins1951stochastic}. And in recent years, SGD has been widely used to minimize the empirical risk in machine learning community~\cite{Shalev-Shwartz:2007:PPE:1273496.1273598,shamir2013stochastic,Bottou2016}. Although the use of stochastic gradient makes SGD has low per iteration cost, a rather high variance introduced by the stochastic gradient will slow down the convergence.

To effectively reduce the variance of stochastic gradient in SGD, some variance reduction methods in statistics, such as importance sampling, control variates and stratified sampling~\cite{Ross2013153,mcbook}, were introduced to stochastic optimization problems. We divide the related work for variance reduction in SGD into two categories:

\begin{enumerate}
  \item From the aspect of optimization variable.
  \item Form the aspect of dataset.
\end{enumerate}

SVRG~\cite{johnson2013accelerating}, SAGA~\cite{NIPS2014_5258} and SDCA~\cite{shalev2013stochastic} can be regarded as the variance reduction methods from the aspect of optimization variable. All of them can reduce the variance to 0 asymptotically. SVRG and SAGA adopt the control variate method to reduce variance. The control variates in both algorithms are the stochastic gradients with respect to the optimization variables of a previous step. SDCA can be regarded as adopting antithetic sampling to reduce variance as explained in~\cite{johnson2013accelerating}. The convergence properties of these methods are different from that of SGD. So, generally, we regard them as different methods.

%

On the other hand, utilizing the internal structure of dataset to proceed variance reduction was considered in~\cite{NIPS2013_5034,zhao2014accelerating,zhao2015stochastic}. In~\cite{NIPS2013_5034}, the control variates were constructed via low-order approximations to the stochastic gradient. \cite{zhao2014accelerating} utilized the clustering information of dataset to divide the whole dataset into clusters and performed the stratified sampling. While~\cite{zhao2015stochastic} studied the importance sampling in SGD, where an nonuniform sampling distribution is constructed according to the internal structure of dataset. \cite{NIPS2017_7025,NIPS2016_6579} developed the importance sampling methods further. Although these methods may not reduce the variance to 0, they can be regarded as complementary to other variance reduction methods.

The above variance reduction methods can be used in combination. \cite{xiao2014proximal,Shen:2016:AVR:3060832.3060899} applied importance sampling to SVRG. \cite{csiba2015stochastic} proposed a variant of SDCA, which also combined the importance sampling methods. \cite{NIPS2016_6403} improved SVRG by exploiting clustering structure inside datasets.

This paper introduces a new method that utilizes the antithetic sampling to reduce the variance in SGD. This method is complementary to the existing variance reduction methods.

\section{SGD with Antithetic Sampling}

\subsection{Problem Setting}
In this paper, we focus on the binary classification task with a set of samples $ \{ (x_1, y_1), (x_2, y_2), \dots, (x_n, y_n) \} $, where each $ x_i \in \mathbb{R}^d $ is a $d$-dimensional instance and $ y_i $ is the corresponding class label assigned to $x_i$. For a sample $ (x_i, y_i) $, the loss function $ f_i(w) $ from $ \mathbb{R}^d $ to $ \mathbb{R} $ is introduced. To learn the classifier, we need to find an approximate solution of the following optimization problem
\begin{equation}\label{eq:1}
  \min\limits_{w \in \mathbb{R}^d } f(w) := \frac{1}{n}\sum\limits_{i=1}^{n} f_i (w)
\end{equation}
Stochastic Gradient Descent~(SGD) is a popular method to solve the above optimization problem, it is an iterative algorithm. At each iteration $ t = 1, 2, \dots $, an index $i_t$ will be uniformly randomly sampled from $ \{1,2,\dots,n\} $, then $w$ will be updated as
\begin{equation}\label{eq:2}
  w_{t+1} = w_t - \eta_t \nabla\! f_{i_t} (w_t),
\end{equation}
where $\eta_t$ is a positive stepsize.

The batch approach is a nature modification. For a mini-batch of indices $ \mathcal{B}_t $ with size $b$, the update of $w$ is
\begin{equation}\label{eq:3}
  w_{t+1} = w_{t} - \frac{\eta_t}{b} \sum\limits_{s \in \mathcal{B}_t}^{} \nabla\! f_s (w_t ),
\end{equation}
where $ \mathcal{B}_t = \{ i_t^{j} \in \{ 1,2,\dots,n \} | j = 1,2,\dots,b \} $ is i.i.d. uniformly sampled with replacement from the set of indices. This sampling process is the distinction to the way in our antithetic sampling.

In the rest of this paper, we'll use SGD to refer to both SGD and mini-batch SGD cases. The stochastic gradients are unified as $g_t$. Then, we can unify (\ref{eq:2}) and (\ref{eq:3}) as:
\begin{equation}\label{eq:2a}
    w_{t+1} = w_t - \eta_t g_t,
\end{equation}

 The expectation of a random variable $X$ is denoted by $\mathbb{E}[X]$. While we use $\mathbb{E}_ t[\cdot]$ to denote an expected value taken with respect to $ \{ i_t^{j} | j = 1,2,\dots,b \} $, the random variable appeared in $t$-th iteration. Note that $g_t$ is an unbiased estimator of the full gradient,
\begin{equation}\label{eq:2aa}
    \mathbb{E}_t [g_t] = \nabla f(w) = \frac{1}{n} \sum\limits_{i=1}^{n} f_i (w).
\end{equation}
In this paper, we'll keep this unbiased property.

\subsection{Impacts of Variance}
In this subsection, we analyse the impacts of variance in SGD from the general perspective. The analysis covers a variety of situations based on different assumptions, such as convex, strongly convex and smoothness.

For the random vector $g_t$, we denote the variance
\begin{equation}\label{eq:4}
  \text{Var}(g_t) = \mathbb{E}_t\! \left\| g_t - \mathbb{E}_t g_t \right\|^2 = \mathbb{E}_t \!\left\| g_t \right\|^2 - \left\|\mathbb{E}_tg_t\right\|^2.
\end{equation}
Since $ \mathbb{E}_t g_t = \nabla f(w_t)$, which is identical under different sampling process, we'll use the second moment $ \mathbb{E}_t\! \left\| g_t\right\|^{\mathsmaller 2} $ to refer to the variance. For the smooth optimization problem, the Lipschitz continuous gradient assumption is usually used to describe the smoothness of the function.

Formally, if $f$ has the Lipschitz continuous gradient with parameter $L$, then for any $ x,y \in \mathbb{R}^d$, we have
    \begin{equation}\label{eq:6}
      f(x) \leq f(y) + \left< \nabla\! f(y), x-y\right> + \frac{L}{2}\left\| x-y \right\|^2.
    \end{equation}
Lipschitz continuous gradient assumption is widely used. It establishes convergence guarantee for stochastic gradient on an assumption of smoothness of the objective function.

By the inequality (\ref{eq:6}), the iterations of SGD satisfy
\begin{eqnarray}
  f(w_{t+1})\! - \! f(w_t) \hspace{-.08in} &  \leq & \hspace{-.08in} \left<\nabla\! f(w_t ), w_{t+1}\! -\! w_t \right> \! + \! \frac{L}{2}\left\|w_{t+1} \! - \! w_t \right\|^2 \nonumber \\
 \hspace{-.08in}  &=& \hspace{-.08in} -\eta_t \left< \nabla\! f(w_t ), g_t \right>  + \frac{L}{2}\eta_t^{2} \left\|g_t \right\|^2. \nonumber
\end{eqnarray}
Taking expectations in these inequalities, we can get
\begin{equation}\label{eq:8}
  \mathbb{E}_tf(w_{t+1}) - f(w_t) \leq -\eta_t \left\| f(w_t)\right\|^2 + \frac{L}{2}\eta_t^{2} \mathbb{E}_t \! \left\|g_t \right\|^2.
\end{equation}

So with bounded variance, under proper stepsize, the decrease of $f$ can be guaranteed in the statistical sense. The smaller the variance is, the larger the stepsize that can be adopted, which will result in a faster decrease of $f$.

Some machine learning applications involve non-smooth loss functions without Lipschitz smooth assumption, for example, the support vector machine with the standard non-smooth hinge loss~\cite{Shalev-Shwartz:2007:PPE:1273496.1273598}. SGD can also work by denoting $g_t$ as a subgradient~\cite{shamir2013stochastic}. According to the iterative process of SGD, we have
\begin{eqnarray} \label{eq:9}
  \hspace{-.08in} & & \hspace{-.08in} \mathbb{E}_t[\left\| w_{t+1} - w^{*} \right\|^2] \nonumber \\
  \hspace{-.08in} &=& \hspace{-.08in} \left\| w_t - w^{*} \right\|^2 - 2\eta_t \mathbb{E}_t\!\left<g_t, w_t - w^{*} \right> + \eta_t^{2} \mathbb{E}_t \! \left\|g_t\right\|^2
\end{eqnarray}
Similar to previous analysis, a smaller variance will allow a faster rate of $w$ toward the optimal point.\footnote{Certainly, this analysis is still valid in the smooth case. Sometimes, for the smooth optimization problems, the analysis concerns about function value rather than $w$, which may not refer to (\ref{eq:9}).}

$f(w)$ or $\| w - w^{*} \|^2$ may not decrease effectively if $\mathbb{E}_t \|g_t\|^2$ is rather big. So in most convergence rate analyses of SGD~\cite{Bottou2016,shamir2013stochastic}, a common assumption is
\begin{equation}\label{eq:9a}
    \mathbb{E}\left\| g_t \right\|^2 \leq G^2, \;\;\; t = 1,2,\dots
\end{equation}
This assumption corresponds to a bounded variance. It is made to ensure the effects of variance can be offset by the decreasing stepsize.

For the convergence rate, what we are concern about is the upper bound of $\mathbb{E}f(w_t) - f(w^{*})$ or $ \mathbb{E}[\left\| w_{t+1} - w^{*} \right\|^2] $. Commonly, (\ref{eq:8}) and (\ref{eq:9}) would appear in the analysis of convergence rate. This makes the upper bound to be related to the variance of stochastic gradient. Generally, the upper bound is proportional to $G^2$. A smaller variance generated in each steps will decrease the upper bound of $\mathbb{E}\!\left\|g_t \right\|^2$. Then $G^2$ will be a smaller value, which will lead to a faster convergence.

\textbf{Remark.} For the optimization problem with a non-smooth regularization term $r$:
\begin{equation}\label{eq:9aa}
  \min\limits_{w \in \mathbb{R}^d } P(w) := \frac{1}{n} \sum\limits_{i=1}^{n} f_i (w) + r(w),
\end{equation}
SGD with proximal operation~(Prox-SGD) can be used to solve it. In this case, the variance of stochastic gradient has a similar influence on convergence rate~\cite{duchi2010composite}. So, the analyses of variance reduction for (\ref{eq:1}) are still applicable.

\subsection{Antithetic Sampling}
In this subsection, we'll introduce the antithetic sampling strategy to reduce the variance.

Without loss of generality, we consider mini-batch SGD with batch size 2. In this case,
\begin{equation}\label{eq:10}
  g_t = \frac{1}{2}\left( \nabla\! f_i (w_t ) + \nabla\! f_j (w_t )\right), \; i, j \in \{ 1,2,\dots,n \}.
\end{equation}
In the traditional setting, $i$ and $j$ are i.i.d., then
\begin{eqnarray}
   \text{Var}(g_t ) \hspace{-.08in}  &=& \hspace{-.08in}  \frac{1}{4} \left(\text{Var}(\nabla\! f_i (w_t)) + \text{Var}(\nabla\! f_j (w_t)) \right) \nonumber\\
  \hspace{-.08in} &=& \hspace{-.08in}  \frac{1}{2}\text{Var}(\nabla\! f_i (w_t ))
\end{eqnarray}

If we adopt a new sampling strategy in which $i$ and $j$ may not independent, but still have identically distribution. Then we have
\begin{equation}\label{eq:11}
  \text{Var}(g_t) = \frac{1}{2} \left\{ \text{Var}(\nabla\! f_i (w_t)) + \text{Cov}(\nabla\! f_i (w_t), \nabla\! f_j (w_t)) \right\},
\end{equation}
where
\begin{eqnarray}
   \hspace{-.08in} & & \hspace{-.08in} \text{Cov}(\nabla\! f_i (w_t), \nabla\! f_j (w_t))  \\
  \hspace{-.08in} &=& \hspace{-.08in} \mathbb{E}_t \! \left\{ \nabla f_i (w_t)^{\mathsmaller T} \nabla f_j (w_t) \right\} \! - \! \mathbb{E}_t [\nabla f_i (w_t)]^{\mathsmaller T}\mathbb{E}_t [\nabla f_j (w_t)] \nonumber
\end{eqnarray}

If $\text{Cov}(\nabla\! f_i (w_t), \nabla\! f_j (w_t))\!<\!0$, which means $\nabla f_j (w_t)$ may be the one that somehow indicates the opposite direction of $\nabla f_i(w_t)$, then we can derive the stochastic gradient $g_t$ with smaller variance. And the closer $\nabla f_j (w_t )$ and $ 2\mathbb{E}_t [g_t] \!- \! \nabla f_i (w_t ) $, the better we can expect to reduce more variance. Such stochastic gradients $\nabla f_i (w_t)$ and $ \nabla f_j (w_t) $ can be regarded as negative correlated. Since $i$ and $j$ have identically distribution, $g_t$ is still an unbiased estimator of full gradient. For the negative correlated $\nabla f_i (w_t)$ and $\nabla f_j (w_t)$, we regard corresponding samples $i$ and $j$ to be antithetic to each other.

\begin{figure}
  \centering
  \includegraphics[width=0.25\textwidth]{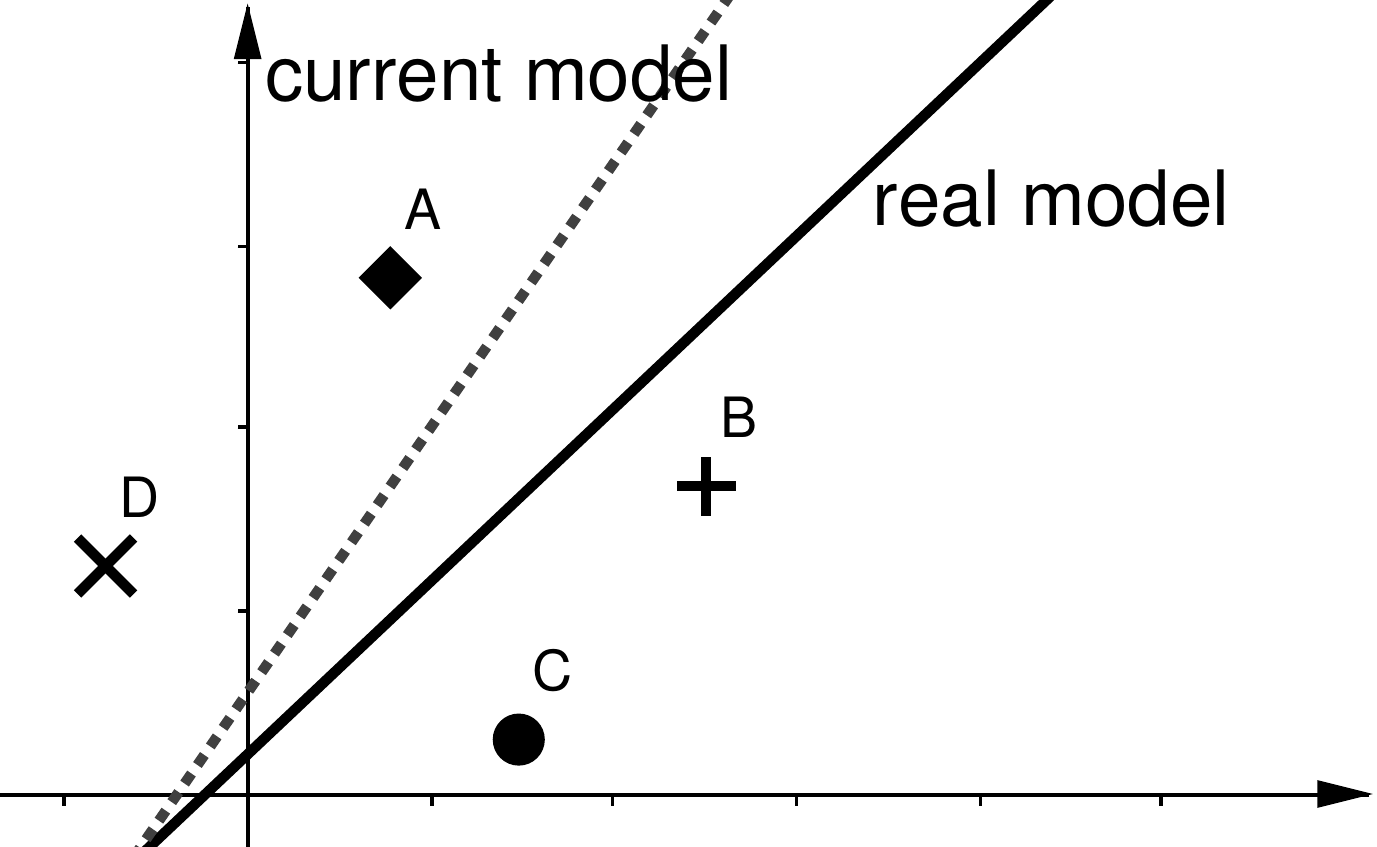}
  \caption{An illustration to explain the effect of antithetic sampling. A, B, C, D are four samples presented, solid line represents the real model we want to get, dotted line represents the model we already got in previous iterations. In the next iteration, suppose we use sample A to train, then the dotted line will tend to A. It is not the direction to the real model as this direction is not obtained through all training samples. If both A and B are selected, then B can rectify the influence of A in some extent, which will make the direction more accurate, then the current model will better approximate the real model. For C, the correction may not as effective as B. However, D will exacerbate the negative impact. So, we can regard B as the antithetic sampling of A in the current iteration.}\label{fig:1}
\end{figure}

In each iteration, it is supposed that we have gotten the pairing information between samples, under which one sample is antithetic to its partner. Then we draw a pair of samples to calculate the mini-batch stochastic gradient. The variance of $g_t$ obtained under this process is smaller than that got under tradition i.i.d. uniformly sampling. As we have explained, a smaller variance corresponds to a smaller $\mathbb{E}_t\!\left\|g_t \right\|^2$, which will directly result in a better convergence performance. Figure~\ref{fig:1} illustrates the effect of antithetic sampling.

\begin{algorithm}[htpb]
\caption{Stochastic Gradient Descent with Antithetic Sampling} \label{alg:SGD-as}
\begin{algorithmic}[1]
    \FOR{$ t = 1,2,\ldots, $}
    \STATE Get Antithetic Table $S$.
    \STATE Uniformly sample $ i_t $ from $ \{1, \dots,n\} $;
    \STATE Get the antithetic sample $j_t = S_{i_t}$;
    \STATE Update $ w_{t + 1} = w_t - \frac{\eta_t}{2}\left( \nabla\! f_{i_t} (w_t ) + \nabla\! f_{j_t} (w_t )\right) $;
    \ENDFOR
\end{algorithmic}
\end{algorithm}

Utilizing above ideas, we propose SGD with antithetic sampling~(SGD-as), which is shown as Algorithm~\ref{alg:SGD-as}. To implement the antithetic sampling, in each step, for each sample $i \in \{1,\dots,n\}$, we need to get its antithetic sample $j$. All the antithetic information is stored in the table $S$. By denoting $S_i = j$, $S$ records the antithesis of all samples. It is a permutation of $\{1,\dots,n\}$ since the unbiased property of mini-batch stochastic gradient. The SGD-as iteration requires the calculation of $S$ at each step, which is clearly inefficient. But for the binary classification, we'll show that $S$ can be calculated beforehand and reused in each iteration. Although Algorithm~\ref{alg:SGD-as} considers the case where the batch size is 2, this algorithm can be generalized to the cases with bigger batch size by simply selecting more pairs of antithetic samples.

\section{Calculate Antithetic Samples}

To conduct the antithetic sampling, we need to calculate the antithetic table. In this section, we'll show this calculation can be efficient in the binary classification problems. We begin with relaxing the calculations of antithetic samples. After that, some common and exact binary classification applications will be discussed in detail.

\subsection{Analysis}

For the stochastic gradient with batch size 2 in SGD, we have
\begin{eqnarray}
  \hspace{-.12in} && \hspace{-.08in}  \mathbb{E}_t\! \left\| g_t \right\|^2 = \frac{1}{4} \mathbb{E}_t\! \left\| \nabla f_i (w_t ) + \nabla f_j (w_t ) \right\|^2 \nonumber \\
  \hspace{-.12in} &=& \hspace{-.1in} \frac{1}{4} \mathbb{E}_t\! \left\{ \left\| \nabla f_i (w_t) \right\|^2 \!+\! \left\| \nabla f_j (w_t) \right\|^2 \!+\! 2 \nabla f_i (w_t)^{\mathsmaller T} \nabla f_j (w_t) \right\}\!. \nonumber
\end{eqnarray}
As only $ \mathbb{E}_t\! \left[\nabla f_i (w_t)^{\mathsmaller T} \nabla f_j (w_t)\right] $ will be affected by the selection of stochastic samples, we just need to consider it.  To minimize $ \mathbb{E}_t\! \left[\nabla f_i (w_t)^{\mathsmaller T} \nabla f_j (w_t)\right] $, we need to calculate the antithetic samples of all samples, which can be represented as the following optimization problem:
\begin{equation}\label{eq:13}
  \! \min\limits_{j_1, j_2, \dots, j_n}\frac{1}{n}  \sum\limits_{i=1}^{n} \nabla f_i (w_t)^{\mathsmaller T} \nabla f_{j_i} (w_t),
\end{equation}
where $j_i$ is the antithetic sample of $i$. Solving this problem requires the calculations of $n$ derivatives in each iteration, which is clearly inefficient. Beyond, this problem itself is difficult to solve. So we seek to reduce $\mathbb{E}_t\!\left\| g_t\right\|^2$ as much as possible rather than minimizing it.

In the traditional SGD setting, $i$ and $j$ are sampled independently, then:
\begin{equation}\label{eq:14}
  \mathbb{E}_t \! \left[\nabla f_i (w_t)^{\mathsmaller T} \nabla f_{j} (w_t)\right]  = \left[ \mathbb{E}_t \! \nabla f_i (w_t)\right]^2 \geq 0,
\end{equation}
So for each pair of $i $ and $j$, if we can guarantee
\begin{equation}\label{eq:15}
  \nabla f_i (w_t)^{\mathsmaller T} \nabla f_j (w_t) < 0,
\end{equation}
then we have $ \mathbb{E}_t \! \left[\nabla f_i (w_t)^{\mathsmaller T} \nabla f_j (w_t)\right] \!<\! 0 $, which will inevitably lead to a smaller $\mathbb{E}_t \!\left\| g_t \right\|^2$ than the traditional setting.

In addition to this ideal case, we can consider relaxing the problem: for an exactly $i$, we aim to sample the $j$ to get a smaller $\nabla f_i (w_t)^{\mathsmaller T} \nabla f_j (w_t)$.

For $\nabla f_i (w_t)^{\mathsmaller T} \nabla f_j (w_t) > 0$, if we can get the upper bound in advance:
\begin{equation}\label{eq:16}
  \nabla f_i (w_t)^{\mathsmaller T} \nabla f_j (w_t) < B_i B_j,
\end{equation}
then we may get a smaller $\nabla f_i (w_t)^{\mathsmaller T} \nabla f_j (w_t)$ by reducing the upper bound. This technique is also used in~\cite{zhao2015stochastic}.

Similarly, for $\nabla f_i (w_t)^{\mathsmaller T} \nabla f_j (w_t) < 0$, if we can get the lower bound in advance:
\begin{equation}\label{eq:17}
    \nabla f_i (w_t)^{\mathsmaller T} \nabla f_j (w_t) > L_i L_j,
\end{equation}
then the value of $\nabla f_i (w_t)^{\mathsmaller T} \nabla f_j (w_t)$ may be reduced with a smaller lower bound.

In the next subsection, we'll show that for common binary classification problems, $B_i, L_i,\, i\in \{1, 2, \dots, n\}$ can be easily obtained and independent of $t$.

\textbf{Remark.} Consider the optimization problem:
\begin{equation}\label{eq:18a}
  \min\limits_{w \in \mathbb{R}^d } f(w) := \frac{1}{n} \sum\limits_{i=1}^{n} f_i (w) = \frac{1}{n}\sum\limits_{i=1}^{n} g_i (w) + h(w),
\end{equation}
where $h(w)$ is the smooth regularization, e.g. $\frac{\lambda}{2} \left\|w \right\|^2 $. Intuitively, since $h(w)$ does not involve $i$, it has no impacts on the variance. So we can follow above analysis without considering $h(w)$. Formally,
\begin{eqnarray}
  \hspace{-.08in} & & \hspace{-.08in} \mathbb{E}_t\! \left[\nabla f_i (w_t)^{\mathsmaller T} \nabla f_j (w_t)\right] \nonumber =  \mathbb{E}_t\! \left[\nabla g_i (w_t)^{\mathsmaller T} \nabla g_j (w_t)\right] \\
  \hspace{-.08in} &&  + 2 \left( \mathbb{E}_t\! \nabla g_i (w) \right)^{\mathsmaller T} \nabla h(w) + \left\| \nabla h(w) \right\|^2 \nonumber
\end{eqnarray}
Only the first term will be affected by the selection of stochastic samples. So we can adopt the above analysis with simply replacing $\nabla f_i(w_t)$ with $\nabla g_i(w_t) $.
%


\subsection{Binary Classification}
In this subsection, we consider two common binary classification applications: logistic regression and support vector machine, one contains smooth loss functions while another contains non-smooth loss functions. We'll show how they correspond to the above analysis.

For the logistic regression problem with $y_i \!\in\! \left\{-1, 1 \right\}, \; i=1,\dots,n$, we need to minimize
\begin{equation}\label{eq:19}
  f(w) = \frac{1}{n} \sum\limits_{i=1}^{n} \log \left( 1 + \exp (-y_i \cdot w^{\mathsmaller '}x_i ) \right).
\end{equation}
Then the stochastic gradient of sample $i$ can be written as
\begin{equation}\label{eq:20}
  \nabla f_i (w) = -\frac{\exp (-y_i \cdot w^{\mathsmaller '} x_i )}{1 + \exp (-y_i \cdot w^{\mathsmaller '} x_i )} \cdot y_i x_i^{\mathsmaller '}.
\end{equation}
Since
\begin{equation}\label{eq:21}
  0 < \frac{\exp (-y_i \cdot w^{\mathsmaller '} x_i )}{1 + \exp (-y_i \cdot w^{\mathsmaller '} x_i )} < 1,
\end{equation}
For the two samples $i$ and $j$, if
\begin{equation}\label{eq:21a}
  y_i y_j x_i^{\mathsmaller '} x_j\! <\! 0,
\end{equation}
then we can get $\nabla f_i (w_t)^{\mathsmaller T} \nabla f_j (w_t)\! < \!0$, which is always hold no matter in which step. Further, if for all $i\in \{1,\dots,n\}$, there exists corresponding $j$ to establish (\ref{eq:21a}), then we can get the mini-batch stochastic gradient with a smaller variance according to the previous analysis,

For the more general case, we have
\begin{equation}\label{eq:21aa}
  0 < \nabla f_i (w_t)^{\mathsmaller T} \nabla f_j (w_t) < y_i y_j x_i^{\mathsmaller '} x_j
\end{equation}
or
\begin{equation}\label{eq:21aaa}
  y_i y_j x_i^{\mathsmaller '} x_j < \nabla f_i (w_t)^{\mathsmaller T} \nabla f_j (w_t) < 0,
\end{equation}
In (\ref{eq:21aa}), $y_i y_j x_i^{\mathsmaller '} x_j$ is the upper bound of $\nabla f_i (w_t)^{\mathsmaller T} \nabla f_j (w_t)$, so $y_ix_i$ here corresponds to $B_i$ in (\ref{eq:16}). While in (\ref{eq:21aaa}), $ y_i y_j x_i^{\mathsmaller '} x_j$ is the lower bound of $\nabla f_i (w_t)^{\mathsmaller T} \nabla f_j (w_t)$, so $y_ix_i$ here corresponds to $L_i$ in (\ref{eq:17}). As a result, we can expect a smaller $\nabla f_i (w_t)^{\mathsmaller T} \nabla f_j (w_t)$ by reducing $y_i y_j x_i^{'} x_j$, and the latter one is independent with $w_t$.



The proposed ideas of calculating antithetic samples are summarized as Algorithm~\ref{alg:anti_table}. The algorithm outputs an antithetic table $S$. And the table will be used in Algorithm~\ref{alg:SGD-as}. For each $i \in \{1,2,\dots,n\}$, $S$ records the corresponding antithetic sample. Note that $S$ just needs to be calculated beforehand, and reused in each iteration.

\begin{algorithm}[htpb]
\caption{Calculate the Antithetic Table $S$ for Binary Classification} \label{alg:anti_table}
\begin{algorithmic}[1]
    \STATE {\bf Input}: $n$ samples $ \left\{ (x_1, y_1),\dots,(x_n, y_n) \right\} $.
    \STATE $ S \leftarrow \{0,0,\dots,0\} $, \;$DB \leftarrow \left\{ (x_1, y_1),\dots,(x_n, y_n) \right\} $
    \FOR{$ i = 1 $ \TO $ n $}
    \STATE $j = \text{NearestNeighborID} ((x_i, y_i), DB, y_i y_j x_i^{\mathsmaller '} x_j) $;
    \STATE $ S[i] = j $, remove $(x_j, y_j)$ from $DB$;
    \ENDFOR
    \STATE {\bf Ouput}: $S$.
\end{algorithmic}
\end{algorithm}

NearestNeighborID$((x_i, y_i), DB, y_iy_jx_ix_j)$ is an oracle that outputs the nearest neighbor's identity of $x$ in the set $D\!B$ by the metric $y_iy_jx_i^{\mathsmaller '}x_j $. Many fast nearest neighbor algorithms can be used by replacing the common distance with the metric we define. To avoid two samples have the same antithesis, $(x_j, y_j)$ is removed from $DB$ in step 5. As a result, $S$ is a permutation of $\{1, \dots, n\}$, $i$ and $j$ have an identical distribution. Generally, due to datasets need to keep positive features, the divergence between $x_i$ and $x_j$ are not large enough to lead to a negative  $x_i^{\mathsmaller '}x_j$. In this case, $y_i$ and $y_j$ play a significant role in determining the sign of  $y_iy_jx_i^{\mathsmaller '}x_j$. So the mutually antithetic samples usually have the opposite labels if possible.


Now, consider the optimization problem IN support vector machine
\begin{equation}\label{eq:23}
  f(w) := \frac{\lambda}{2}\left\| w\right\|^2 + \frac{1}{n}\sum\limits_{i=1}^{n} \max\!\left\{ 0, 1-y_i \cdot w^{\mathsmaller '}x_i \right\},
\end{equation}
which satisfies the form in (\ref{eq:18a}). So, we just need to consider the term which involves the stochastic sample. For $ (x_i, y_i) $, the (sub-)gradient is:
\begin{equation}\label{eq:25}
  \nabla g_i (w) = - \mathbbm{1}\!\left\{ y_i \cdot w^{\mathsmaller '}x_i \leq 1 \right\} \cdot y_i x_i.
\end{equation}
As $ \mathbbm{1}\!\left\{ y_j \cdot w^{\mathsmaller '}x_j \leq 1 \right\}\! \in\! \{0, 1\} $, we have
\begin{equation}\label{eq:26}
  0 \leq \nabla g_i (w_t)^{\mathsmaller T} \nabla g_j (w_t) \leq y_i y_j x_i^{\mathsmaller '} x_j
\end{equation}
or
\begin{equation}\label{eq:27}
  y_i y_j x_i^{\mathsmaller '} x_j \leq \nabla g_i (w_t)^{\mathsmaller T} \nabla g_j (w_t) \leq 0,
\end{equation}
which is consistent with the analysis in logistic regression. So Algorithm~\ref{alg:anti_table} is still available for the optimization problem in support vector machine.

\textbf{Remark. }For the logistic regression with $y_i \in \{0, 1\}$. We would like to minimize the loss function:
\begin{equation}\label{eq:32}
  f(w) = \frac{1}{n} \sum\limits_{i=1}^{n} \left[ y_i \log\mu_i + (1-y_i) \log(1-\mu_i)  \right].
\end{equation}
where $\mu_i = \frac{1}{1 + \exp (- w^{\mathsmaller '}x_i )}$. The form is different from (\ref{eq:19}). For two samples $i$ and $j$,
\begin{equation}\label{eq:33}
  \nabla f_i (w) \nabla f_j (w) = (\mu_i - y_i)(\mu_j - y_j)x_i^{\mathsmaller '}x_j.
\end{equation}
Denote $c_{ij} = \left|(\mu_i - y_i)(\mu_j - y_j)\right|$, then
\begin{equation}\label{eq:34}
  \left\{
    \begin{aligned}
        \nabla f_i (w)^{\mathsmaller T} \nabla f_j (w)  &= - c_{ij}x_i^{\mathsmaller '}x_j \;\;\; \text{if}\; y_i \neq y_j \\
        \nabla f_i (w)^{\mathsmaller T} \nabla f_j (w)  &=  c_{ij}x_i^{\mathsmaller '}x_j \;\;\; \text{if}\; y_i = y_j
    \end{aligned}
  \right.
\end{equation}
By denoting $z_i = 2y_i-1$ such that $z_i \in \{-1, 1\}$, (\ref{eq:34}) is equal to
\begin{equation}\label{eq:35}
    \nabla f_i (w)^{\mathsmaller T} \nabla f_j (w) = c_{ij} z_i z_j x_i^{\mathsmaller '} x_j.
\end{equation}
Note that $0\!<\!c_{ij}\!<\!1$, $z_iz_jx_i^{\mathsmaller '}x_j$ is the upper bound or the lower bound of $\nabla f_i (w)^{\mathsmaller T} \nabla f_j (w)$, which is consistent with the former analysis. Algorithm~\ref{alg:anti_table} still work for this problem by simply replacing $y_i$ with $z_i$ for each $i \in \{1,\dots,n\}$ in this algorithm.

\section{Experiments}
In this section, we evaluate the empirical performance of the proposed algorithm by comparing it to SGD with uniform sampling.

\begin{table}
  \centering
  \caption{Datasets used in experiments}\label{table:1}
  \begin{tabular}{|c|c|c|c|c|}
      \hline
      Dataset &  Size & $\#$ features & $\lambda$ \\
       \hline
       \hline
      Sonar & 208 & 60 & $ 10^{-2} $ \\
      breast-cancer & 683 & 10 & $10^{-2}$ \\
      splice & 1000 & 60 & $10^{-3}$ \\
      ijcnn & 35000 & 22 & $10^{-4}$ \\
      fourclass & 862 & 2 & $ 10^{-2} $ \\
      diabetes & 768 & 8 & $ 10^{-2} $ \\
      \hline
  \end{tabular}
\end{table}

In the experiments, we consider two optimization problems in binary classification: $l_2$-regularized logistic regression and SVM. One is a smooth optimization problem, while another is a non-smooth optimization problem. Both of them have a $l_2$ regularization term $\frac{\lambda}{2}\left\|w \right\|^2$. The experiments were performed on several real world datasets downloaded from the LIBSVM website\footnote{\url{https://www.csie.ntu.edu.tw/~cjlin/libsvmtools/}}. The dataset characteristics and regularization parameters $ \lambda $ are provided in the Table~\ref{table:1}.

To make a fair comparison, both algorithms adopt the same setup in our experiments. The size of mini-batch is 2. The initial value of $x$ is set to $0$. In the $t$-th iteration, the step size is set as $ \eta_t = \frac{\eta_0}{1 + \eta_0 \cdot \eta \cdot t} $, where $\eta_0$ is the initial step size and $\eta>0$.

\begin{figure}
  \centering
  \includegraphics[width=0.43\textwidth]{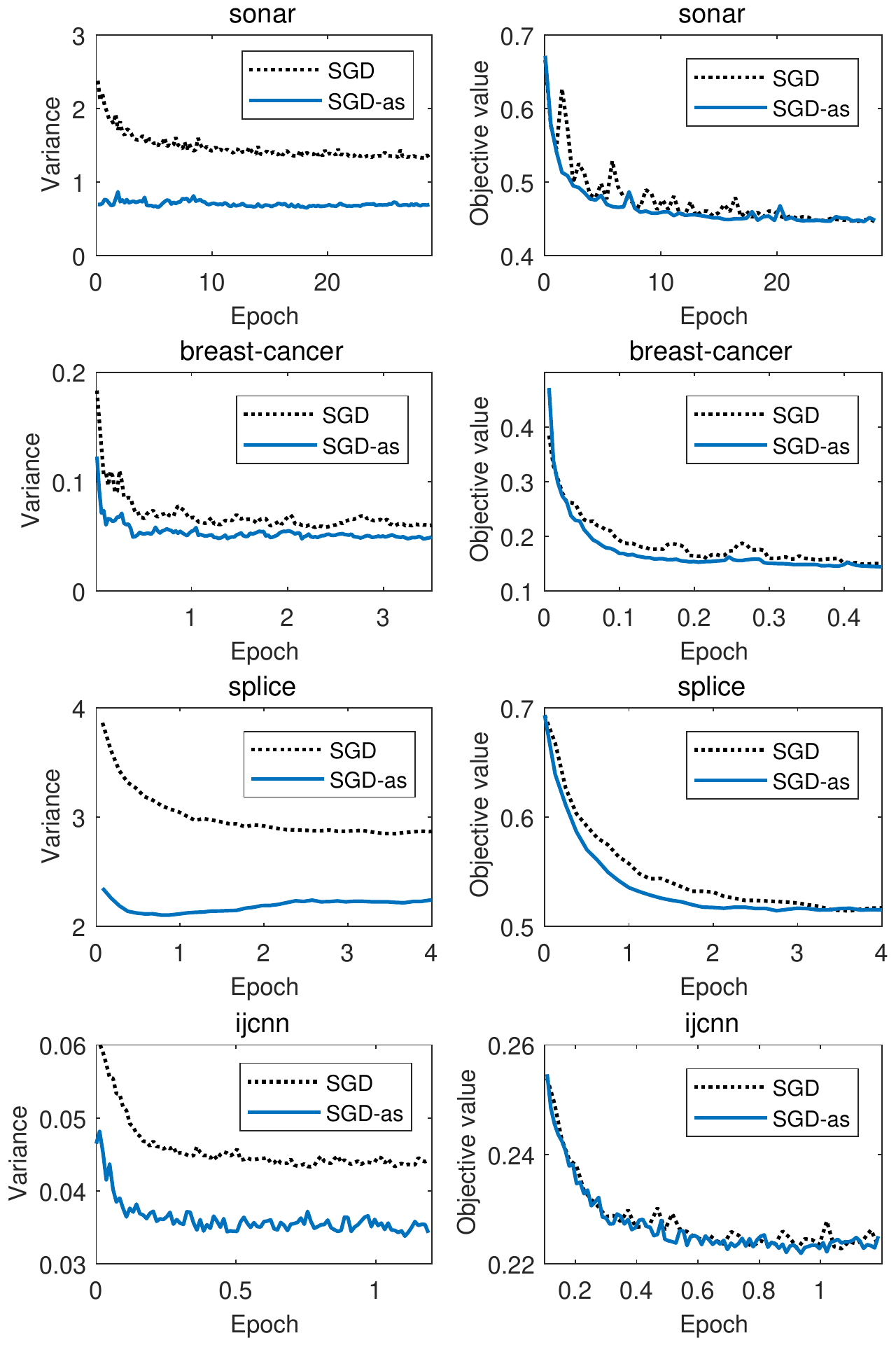}
  \caption{Comparison between SGD-as with SGD in logistic regression}\label{fig:2}
\end{figure}

We report the variances of the stochastic gradient in these two algorithms to check whether antithetic sampling can effectively reduce the variance, and evaluate the algorithms' performance by objective value $f(w_t)$ on training data. By comparing variance with the objective value, we can observe the impacts of variance on the algorithms' performance.

\begin{figure}
  \centering
  \includegraphics[width=0.43\textwidth]{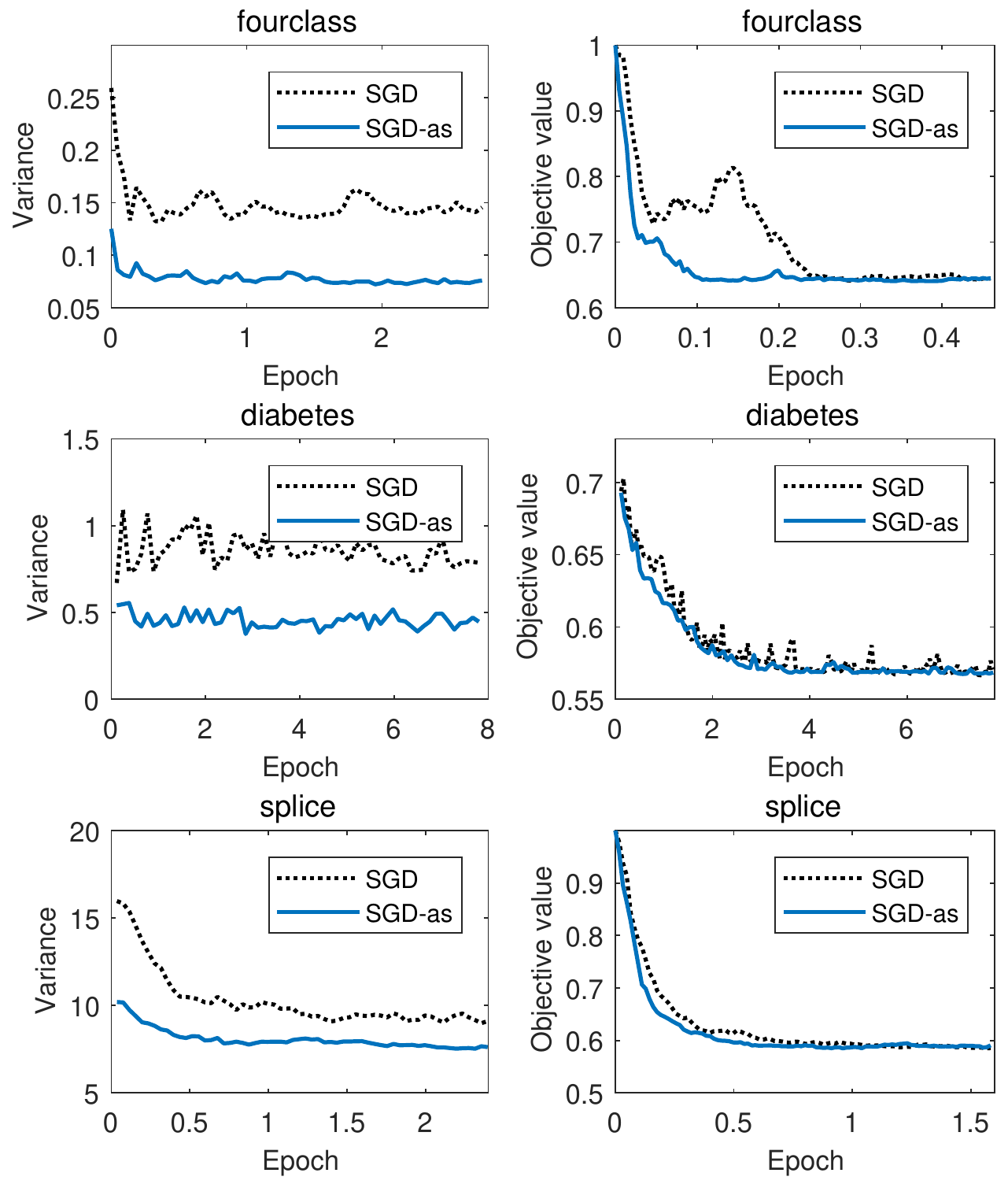}
  \caption{Comparison between SGD-as with SGD in SVM.}\label{fig:3}
\end{figure}

Figure \ref{fig:2} summarizes the experiment results on logistic regression, and Figure \ref{fig:3} summarizes the results on SVM. According to the left column in each figures, we can observe that the variance of SGD-as is significantly smaller than that of SGD with uniform sampling, especially in the beginning of the iterations. This demonstrates the effectiveness of the proposed strategy to reduce the variance of the unbiased stochastic gradient estimators. In each figure, the right column shows the primal objective value of SGD-as in comparison to that of SGD with uniform sampling. We can observe that SGD-as converges faster and is much more stable than SGD. Since both algorithms iterate from $w=0$ and adopt the same step size in each iteration, this observation again implies that our antithetic sampling stratified sampling strategy is more effective to reduce the variance of stochastic gradient. For a certain dataset, we can see as the number of iterations increases, although the variance remains stable, the jitter of the convergence curve is slowing down. This is due to the decreasing step size weakens the impacts of variance. This is also the reason why SGD with the decreasing step size can converge in the presence of variance.

\section{Conclusion}
This paper studied antithetic sampling to reduce the variance for Stochastic Gradient Descent method. We considered the specific binary classification problems in detail. For some common binary classification problems, such as logistic regression and SVM, we proposed the fixed antithetic sampling strategy, which is practical and efficient. Promising empirical results validated the effectiveness of our technique.


\appendix

\bibliographystyle{named}
\bibliography{ijcai18}

\end{document}